\documentclass{article} 
\usepackage{iclr2026_conference,times}

\usepackage{amsmath,amsfonts,bm}

\def\eqref#1{equation~\ref{#1}}

\def\1{\bm{1}}

\DeclareMathAlphabet{\mathsfit}{\encodingdefault}{\sfdefault}{m}{sl}
\SetMathAlphabet{\mathsfit}{bold}{\encodingdefault}{\sfdefault}{bx}{n}

\usepackage{hyperref}
\usepackage{url}
\usepackage{graphicx}
\usepackage{algorithm}
\usepackage{algorithmic}

\title{CHLU: The Causal Hamiltonian Learning Unit as a Symplectic Primitive for Deep Learning}

\author{Pratik Jawahar\\
University of Manchester\\
\texttt{pratik.jawahar@cern.ch} \\
\And
Maurizio Pierini \\
CERN \\
}

\iclrfinalcopy 
\begin{document}

\maketitle

\begin{abstract}
Current deep learning primitives dealing with temporal dynamics suffer from a fundamental dichotomy: they are either discrete and unstable (LSTMs) \citep{pascanu_difficulty_2013}, leading to exploding or vanishing gradients; or they are continuous and dissipative (Neural ODEs) \citep{dupont_augmented_2019}, which destroy information over time to ensure stability. We propose the \textbf{Causal Hamiltonian Learning Unit} (pronounced: \textit{clue}), a novel Physics-grounded computational learning primitive. By enforcing a Relativistic Hamiltonian structure and utilizing symplectic integration, a CHLU strictly conserves phase-space volume, as an attempt to solve the memory-stability trade-off. We show that the CHLU is designed for infinite-horizon stability, as well as controllable noise filtering. We then demonstrate a CHLU's generative ability using the MNIST dataset as a proof-of-principle.
\end{abstract}

\section{Introduction}
The modeling of physical systems and long-term dependencies remains an open challenge in deep learning. Deep learning architectures, while empirically successful, fail to learn implicit conservation laws that govern physical reality~\citep{vafa_what_2025}. Discrete units (eg. RNNs~\citep{salehinejad_recent_2018}, Transformers~\citep{vaswani_attention_2017}) often lack interpretable state dynamics, while continuous-depth models like Neural ODEs (NODEs)~\citep{chen_neural_2019} typically model dissipative systems, making them unsuitable for long-term information preservation. Hamiltonian Neural Networks~\citep{greydanus_hamiltonian_2019} have emerged to enforce energy conservation in traditional Neural Networks, but they are predominantly designed for \textit{simulation} rather than \textit{general inference} on high-dimensional data.

We introduce the \textbf{Causal Hamiltonian Learning Unit (CHLU)}\{pronounced: \textit{clue}\}, a computational learning primitive that treats energy conservation not as a target to be learned, but as a structural prior to moderate latent evolution. A CHLU integrates relativistic mechanics~\citep{hawking_general_2010} with symplectic geometry~\citep{siegel_symplectic_2014} and redefines information propagation as an evolution of the unit's internal state. This work is an early stage description of a CHLU and a comprehensive study is left for future work. Our contributions are two-fold:

\paragraph{Relativistic Kinetic Governor:} We introduce a configurable speed limit $c$, that ensures bounded velocity updates inherently, acting as a structural constraint on kinetic stability, which we intend to utilize in future work on networks of multiple CHLUs. In contrast, symplectic integrators are incorporated in RNNs~\citep{chen_symplectic_2020, erichson_lipschitz_2021, salem_recurrent_2022}, specifically to solve vanishing gradients. Other works like, LorentzNet~\citep{gong_efficient_2022} have explored relativistic symmetries in high-energy physics data, while a CHLU's relativistic definitions target temporal stability of state updates. We reserve the extension to a relativistic attention mechanism for future work.

\paragraph{Action Propagation:} We use a thermodynamic modification of the Wake-Sleep algorithm~\citep{hinton_wake-sleep_1995}, optimizing the \textit{action} difference between ``clamped-wake'' and ``free-sleep'' trajectories. Standard Energy-Based Models (EBMs), like Helmholtz Machines~\citep{dayan_helmholtz_1995} and Restricted Boltzmann Machines~\citep{smolensky_information_1986}, use Contrastive Divergence~\citep{carreira-perpinan_contrastive_2005} or Equilibrium Propagation~\citep{scellier_equilibrium_2017} to derive gradients by relaxing a network to a static fixed point. A CHLU's action propagation is a generalization of these methods. On the other hand, unlike Denoising Hamiltonian Networks~\citep{deng_denoising_2025} which combine dynamics with diffusion masking, a CHLU treats generation as thermodynamic relaxation on a learned potential energy surface.

\section{Methodology: The Geometry of a CHLU}

\subsection{The Separable Hamiltonian Engine}
The core of a CHLU is a dynamical system defined by a learnable Hamiltonian function $\mathcal{H}(q, p)$. Unlike standard Euclidean Hamiltonians, we utilize a \textbf{Relativistic Hamiltonian} to strictly impose causal evolution bounds and ensure energy stability.

The latent state $\mathbf{z} = (\mathbf{q}, \mathbf{p})$, represents generalized positions and momenta. The Hamiltonian $\mathcal{H}$ is:

\begin{equation}
    \mathcal{H}(q,p) = \underbrace{T(\mathbf{p})}_{\text{Relativistic Kinetic Governor}} + \underbrace{V_\theta(q)}_{\text{Learnable Potential Energy}} + \underbrace{\alpha \|q\|^2}_{\text{Global Confinement Potential}}
\end{equation}
where $V_\theta(q)$ is a learnable non-linear potential energy function parameterized by a neural network. The final term is a weak quadratic confinement potential parameterized by $\alpha$, which acts as a regularizer when $V_\theta$ is flat (analogous to a weak, local gravitational potential).

The state $z = (q, p)$ evolves according to Hamilton's equations:
\begin{equation}
    \frac{dq}{dt} = \frac{\partial \mathcal{H}}{\partial p}, \quad \frac{dp}{dt} = - \frac{\partial \mathcal{H}}{\partial q}
\end{equation}

\subsection{Relativistic Kinetic Governor}
In Newtonian mechanics, the kinetic energy is typically $T(\mathbf{p}) = \frac{1}{2}\mathbf{p}^T \mathbf{M}^{-1} \mathbf{p}$, which allows for unbounded velocities. To prevent such kinetic instabilities we replace this with a Relativistic Kinetic Governor. This component is parameterized by a learnable diagonal, positive-definite mass matrix $\mathbf{M}$, and the hyperparameters, rest mass $m_0$ and the speed of causality $c$. Thus, $T(\mathbf{p})$ is defined as:
\begin{equation}
    T(\mathbf{p}) = \sqrt{c^2 \mathbf{p}^T \mathbf{M}^{-1} \mathbf{p} + m_0^2 c^4}
\end{equation}
This formulation ensures that the velocity $\dot{\mathbf{q}} = \nabla_p T$ saturates at the speed limit $c$ (shaped by the geometry of $\mathbf{M}$) as momentum increases (see App.\ref{sec:limit derivation}), thereby enforcing strictly bounded velocity updates, preventing the kinetic explosions common in unconstrained recurrent architectures.

\subsection{Symplectic Integration via Velocity Verlet}
To ensure the system conserves energy over infinite horizons (or dissipates it controllably), we embed a \textbf{Dissipative Velocity Verlet} integrator~\citep{martys_velocity_1999} directly into the forward pass. This allows us to switch between conservative dynamics ($\gamma=0$) and dissipative convergence ($\gamma > 0$), where $\gamma$ is a friction-parameter incorporated to the step update as:
\begin{align}
    \mathbf{p}_{t+0.5} &= \mathbf{p}_t - \frac{\epsilon}{2} \nabla V_\theta(\mathbf{q}_t) \\
    \mathbf{q}_{t+1} &= \mathbf{q}_t + \epsilon \nabla T(\mathbf{p}_{t+0.5}) \\
    \mathbf{p}^*_{t+1} &= \mathbf{p}_{t+0.5} - \frac{\epsilon}{2} \nabla V_\theta(\mathbf{q}_{t+1}) \\
    \mathbf{p}_{t+1} &= (1-\gamma) \mathbf{p}^*_{t+1} \quad \text{(Dissipative Step)}
\end{align}

\section{Training Dynamics: Hamiltonian Contrastive Divergence}
A CHLU's Energy-based Wake-Sleep algorithm consists of:
\begin{itemize}
    \item \textbf{Wake Phase (Supervised):} The system is tasked with minimizing the Mean Squared Error (MSE) between the prediction (integrator step) and the target. To enforce training stability, we apply a regularization term $\mathcal{L}_{reg}$ penalizing the Lyapunov exponents of the Jacobian.
    \item \textbf{Sleep Phase (Unsupervised):} The system evolves freely from a replay buffer. We update weights to raise the energy of hallucinations (unless they match the data distribution).
\end{itemize}

\begin{algorithm}[h]
\caption{CHLU Wake-Sleep Training (Single Step)}
\label{alg:chlu_training}
\begin{algorithmic}[1]
\REQUIRE Data trajectory $\mathbf{z}_{target}$, Initial state $\mathbf{z}_0$, Learning rate  $\eta$ \newline
\textbf{Wake Phase (Clamped Evolution)}
\STATE $\mathbf{z}_{wake} \leftarrow \text{VelocityVerlet}(\mathbf{z}_0, \nabla \mathcal{H}, \gamma=0)$
\STATE $\mathcal{L}_{wake} \leftarrow \text{MSE}(\mathbf{z}_{wake}, \mathbf{z}_{target}) + \lambda \mathcal{L}_{reg}(\text{Lyapunov})$ \newline
\textbf{Sleep Phase (Free Evolution)}
\STATE $\mathbf{z}_{sleep} \leftarrow \text{Sample}(\text{ReplayBuffer})$
\STATE $\mathbf{z}_{hallucination} \leftarrow \text{VelocityVerlet}(\mathbf{z}_{sleep}, \nabla \mathcal{H}, \gamma=0)$ \newline
\textbf{Action Update}
\STATE $\Delta \theta \propto -\nabla_\theta \mathcal{H}(\mathbf{z}_{wake}) + \nabla_\theta \mathcal{H}(\mathbf{z}_{hallucination})$
\STATE $\theta \leftarrow \theta - \eta \Delta \theta$
\STATE $\text{ReplayBuffer.add}(\mathbf{z}_{hallucination})$
\end{algorithmic}
\end{algorithm}

\noindent The final weight update is proportional to the difference of these two signals:
\begin{equation}
    \Delta \theta \propto -\nabla_\theta \mathcal{H}(z_{wake}) + \nabla_\theta \mathcal{H}(z_{sleep})
\end{equation}
This contrastive signal ensures that the system learns to differentiate between physical signals (low energy) and noise (high energy). This training mechanism is currently an empirical choice. 

\paragraph{Hamiltonian Annealing:}
While the core unit is conservative, we employ the controllable friction coefficient $\gamma$ during inference. Setting $\gamma > 0$ drains entropy from the system, collapsing the state into the nearest stable attractor (ground state) defined by $V_\theta$.

\paragraph{Generative Sampling via Langevin Dynamics:}
To transition from deterministic inference to generative modeling, we couple the Hamiltonian system with stochastic Langevin Dynamics~\citep{brooks_handbook_2011}. During inference, we modify the momentum update as:
\begin{equation}
    d\mathbf{p} = -\nabla V_\theta(\mathbf{q})dt - \gamma \mathbf{p} dt + \sqrt{2\gamma k_B \mathcal{T}} d\mathbf{W}
\end{equation}
where $\gamma$ is the friction coefficient, $\mathcal{T}$ is a temperature hyperparameter, and $d\mathbf{W}$ is a Wiener process (standard Brownian motion). This turns the CHLU into a sampler for the Boltzmann distribution $P(\mathbf{q}, \mathbf{p}) \propto \exp(-\mathcal{H}/k_B \mathcal{T})$. By annealing the temperature $\mathcal{T}$, the system settles into low-energy modes of the learned potential $V_\theta(\mathbf{q})$, effectively "crystallizing" noise into structured data samples.

\section{Scope of Experiments and Results}

For this work we provide two experiments that demonstrate specific inductive biases of a CHLU, followed by an \texttt{MNIST}~\citep{yann_mnist_2010} generation experiment. We choose an LSTM~\citep{hochreiter_long_1997} and a NODE as baselines not to compare performance, but rather to contrast the inductive biases of each algorithm and their learned topologies. We do not tune hyperparameters for any model. See appendices for further plots and discussions. We keep performance comparison experiments and studies on failure modes of a CHLU for a larger, comprehensive future study.

\paragraph{Experiment I: Long-Horizon Stability (Lemniscate Tracing)} We train the three models to learn the trajectory of a Lemniscate, a self-intersecting orbit. Each model is trained on $3$ cycles and asked to infer the shape for $50$ cycles.

\begin{figure}[ht]
\begin{center}
\includegraphics[width=\linewidth]{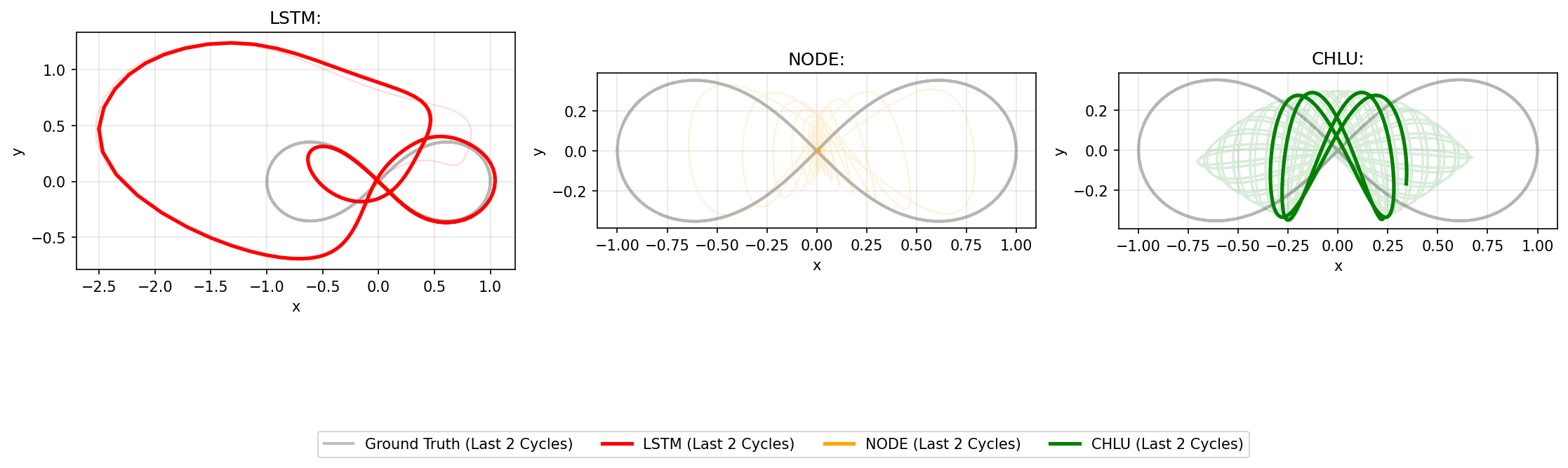}
\end{center}
\caption{Lemniscate tracing experiment. The transparent lines show inference cycles $0-48$ and the solid lines show the final $2$ cycles, for LSTM, NODE and CHLU (left to right)}
\label{fig:exp1_stability}
\end{figure}

The LSTM learns the general shape but over time, numerical errors accumulate positively, pushing the trajectory into a high-energy limit cycle.
The NODE accurately captures short-term dynamics with smooth curves, However, the dissipative nature of NODEs causes the trajectory to spiral inward, eventually collapsing to the origin. The CHLU preserves the shadow Hamiltonian exactly. Despite the learned trajectory being imperfect, the error is bounded. The orbit remains closed and stable indefinitely, demonstrating that symplectic constraints are necessary for long-term topological fidelity.

\paragraph{Experiment II: Kinetic Safety (The Perturbed Sine Wave)}
The models are trained on $100$ sine wave trajectories of length $T=1000$, with frequencies $\omega \sim U(0.5, 2.0)$, where each algorithm outputs the states $(q,p)$. At inference, the initial states are perturbed and the subsequent trajectory is predicted.

\begin{figure}[ht]
\begin{center}
\includegraphics[width=0.7\linewidth]{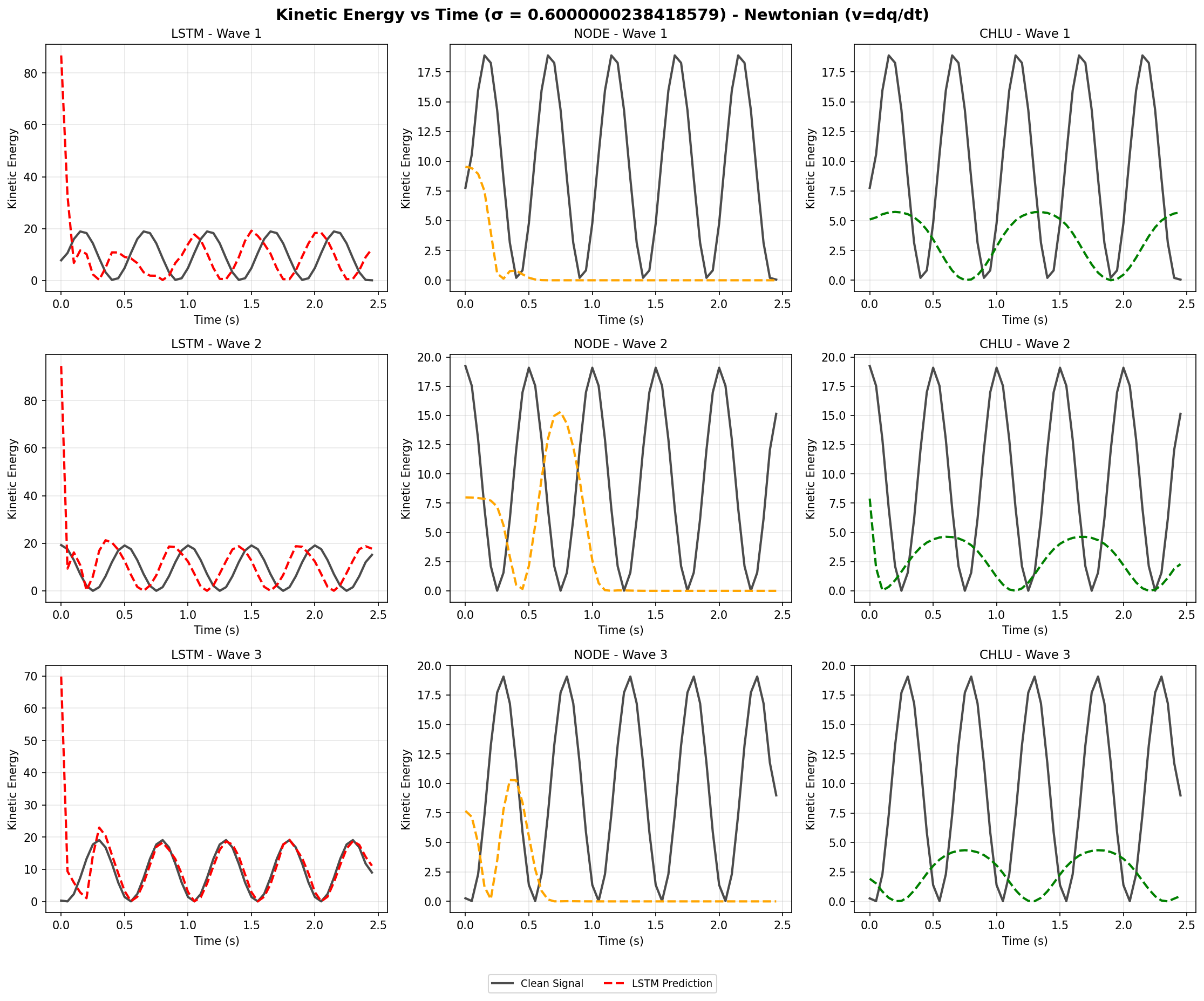}
\end{center}
\caption{The LSTM, NODE and CHLU (left to right), each predicts the state $q$ given the same initial perturbed states, from which we calculate the KE and overlay it on the expected KE.}
\label{fig:exp2_noise}
\end{figure}

The LSTM predicts an initial velocity spike to correct for the noise instantly. In physical systems, such instant, near-infinite acceleration is non-physical and represents a catastrophic failure of the model's internal physics engine. The NODE collapses the wave completely as a trivial solution to MSE reduction. The CHLU, by virtue of the dissipative verlet step, smoothly saturates the velocity at $c$. The perturbation is converted into a phase shift rather than a magnitude divergence. This demonstrates that imposing a causal speed limit is a robust defense against initialization instabilities.

\paragraph{Experiment III: Thermodynamic Generation}
We train a CHLU on $10$k \texttt{MNIST} images. Then we use Langevin Dynamics during the generative phase to produce digits from a starting point that is the centroid of $2$k held out images with added Gaussian noise.

\begin{figure}[ht]
\begin{center}
\includegraphics[width=0.8\linewidth]{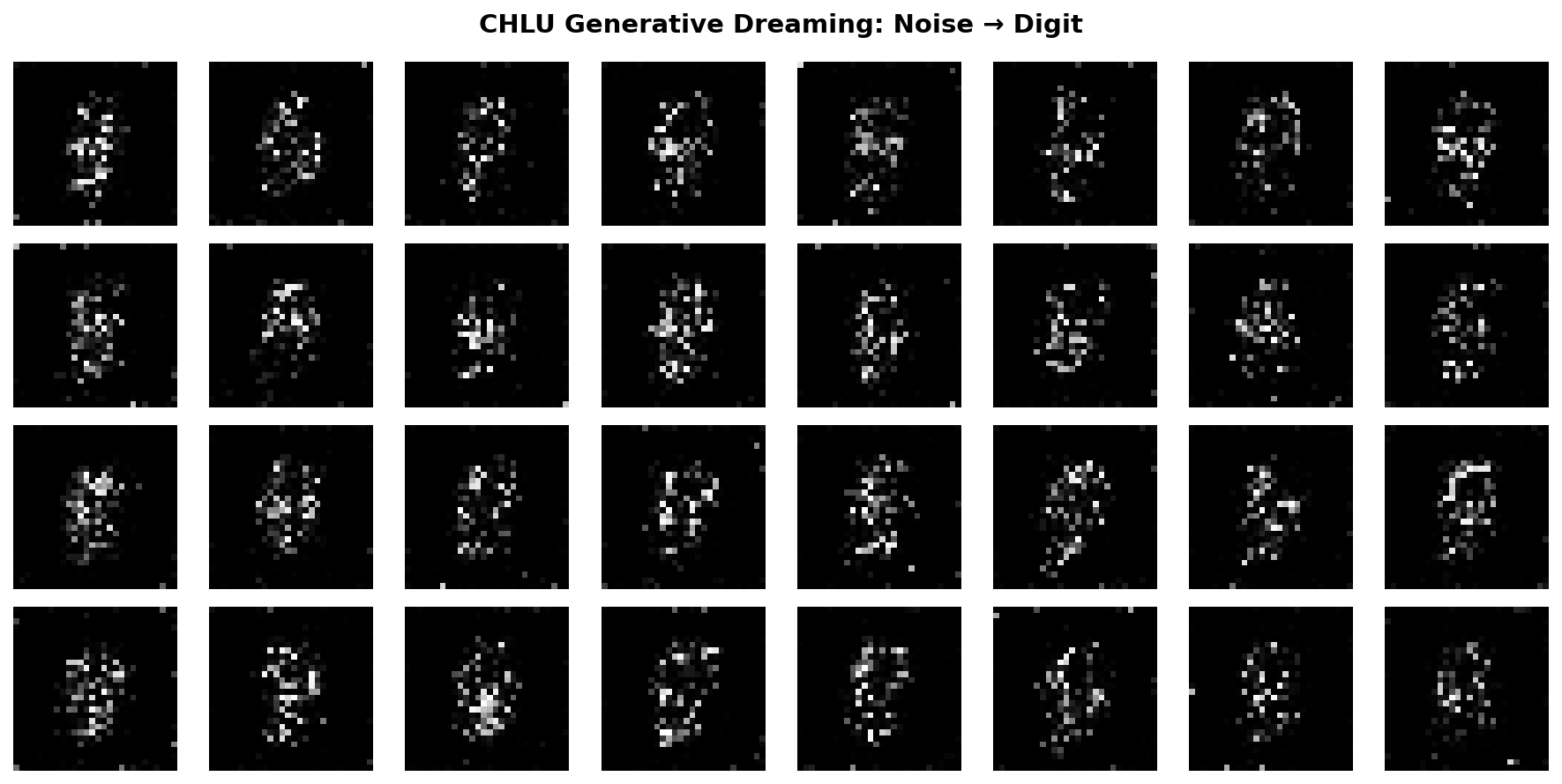}
\end{center}
\caption{\texttt{MNIST} digits generated from the centroid of the test set with added noise.}
\label{fig:exp3_mnist}
\end{figure}

By treating data points as low-energy wells in $V_\theta(q)$, the CHLU generates samples by annealing the input noise. We observe distinct digit modes, although some digits ($3, 5, 8, 9$) are generated more often than others. See Appendices for further results from different generative modes.

\section{Conclusion: From Learning Dynamics to Enforcing Geometry}

Current approaches to temporal modeling suffer from a fundamental trade-off: discrete recurrences (LSTMs) offer expressivity at the cost of stability, while continuous approximations (Neural ODEs) achieve smoothness at the cost of dissipation. We have introduced a CHLU to resolve this dichotomy not by learning better approximation functions, but by enforcing a stricter geometric reality. Our experiments demonstrate that this inductive bias yields distinct, verifiable advantages: topological fidelity, kinetic safety and thermodynamic generation. We envision this primitive serving as a unit in deep, physically consistent networks targeting systems that learn better world models. See Appendices for discussions on future directions. Our software package is available at~\citep{jawahar_praktikal24chlu_2026}.

\bibliography{references}
\bibliographystyle{iclr2026_conference}

\newpage
\appendix

\section{Extended Related Work}

\subsection{Theoretical Foundations: Causal Geometries}
Our architectural constraints draw on the holographic principle and the geometry of causal diamonds~\citep{gibbons_geometry_2007}. Unlike standard recurrent units which allow arbitrary information flow, a CHLU restricts latent updates to a strict, learnable light-cone parametrized by $c$. This ensures that the learned dependencies respect causality, stabilizing the learning of long-range temporal dependencies. We argue that this approach with ad-hoc non-causal data masks is as important as an inherently non-causal algorithm with causal data masks (for eg. transformers trained on causally masked natural language data).

\subsection{Advanced Stability: Comparisons to LyTimeT}
Recent works in robust state discovery, such as LyTimeT~\citep{yu_lytimet_2025}, utilize Lyapunov regularization terms in the loss function to constrain otherwise unstable latent dynamics. In contrast, CHLU decouples stability from optimization. The Relativistic Kinetic Governor guarantees global boundedness (BIBO stability) by architectural construction, rendering the system immune to divergence. We utilize Lyapunov regularization strictly for topological sculpting of the potential energy surface $V_\theta(q)$ to ensure that valid data modes form deep, asymptotic attractors, rather than to prevent non-physical explosions.

\subsection{Generative Lineage: RBMs and Helmholtz Machines}
A CHLU shares the thermodynamic DNA of Restricted Boltzmann Machines (RBMs) and Helmholtz Machines, utilizing a ``Wake-Sleep'' contrastive learning objective. However, where RBMs rely on stochastic Gibbs sampling over static bipartite graphs, a CHLU employs deterministic Symplectic Integration over a continuous phase space. Furthermore, unlike Helmholtz machines which require separate recognition and generative weights, a CHLU unifies inference and generation into a single reversible Hamiltonian operator. Because Hamiltonian dynamics are time-reversible, a CHLU's ``recognition'' model is strictly the time-reversal of the ``generative'' model, ensuring geometric consistency between the latent prior and the data likelihood.

\section{Derivation of Relativistic Gradients}
\label{sec:limit derivation}
The canonical momentum update requires the gradient of the relativistic kinetic energy $T(\mathbf{p}) = \sqrt{c^2 \mathbf{p}^T \mathbf{M}^{-1} \mathbf{p} + m_0^2 c^4}$.
The gradient with respect to momentum $\mathbf{p}$ is derived as:
\begin{equation}
    \frac{\partial T}{\partial \mathbf{p}} = \frac{1}{2\sqrt{c^2 \mathbf{p}^T \mathbf{M}^{-1} \mathbf{p} + m_0^2 c^4}} \cdot (2 c^2 \mathbf{M}^{-1} \mathbf{p}) = \frac{c^2 \mathbf{M}^{-1} \mathbf{p}}{\sqrt{c^2 \mathbf{p}^T \mathbf{M}^{-1} \mathbf{p} + m_0^2 c^4}}
\end{equation}
As $|\mathbf{p}| \to \infty$, the term $m_0^2 c^4$ becomes negligible compared to the momentum term. The velocity saturates as:
\begin{equation}
    \lim_{|\mathbf{p}| \to \infty} \nabla_p T \approx \frac{c^2 \mathbf{M}^{-1} \mathbf{p}}{\sqrt{c^2 \mathbf{p}^T \mathbf{M}^{-1} \mathbf{p}}} = c \frac{\mathbf{M}^{-1} \mathbf{p}}{\sqrt{\mathbf{p}^T \mathbf{M}^{-1} \mathbf{p}}}
\end{equation}
This confirms that the velocity is bounded by $c$ within the geometry defined by $\mathbf{M}$.

\section{Further Experimental Results}
The full list of hyerparameters used for the paper are provided as default configuration options with our software package released with this work.

\subsection{Lemniscate Experiment}
\begin{figure}[ht]
\begin{center}
\includegraphics[width=0.8\linewidth]{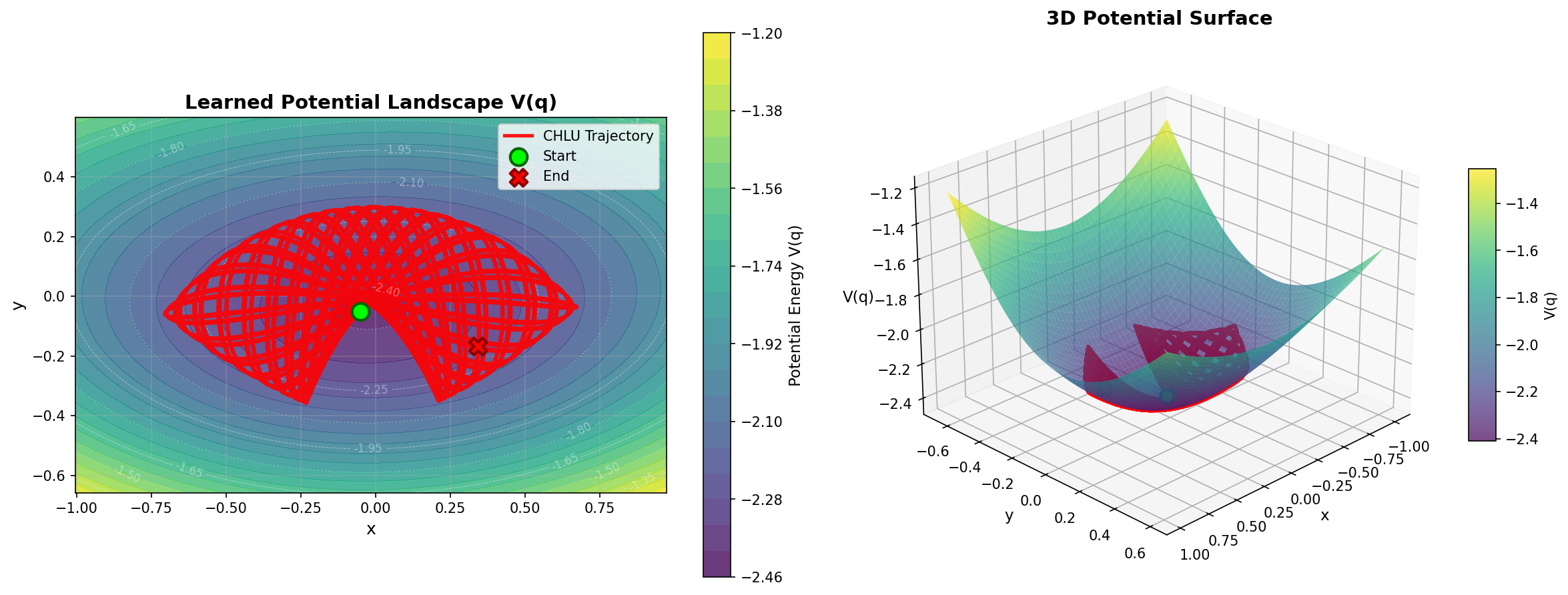}
\includegraphics[width=0.8\linewidth]{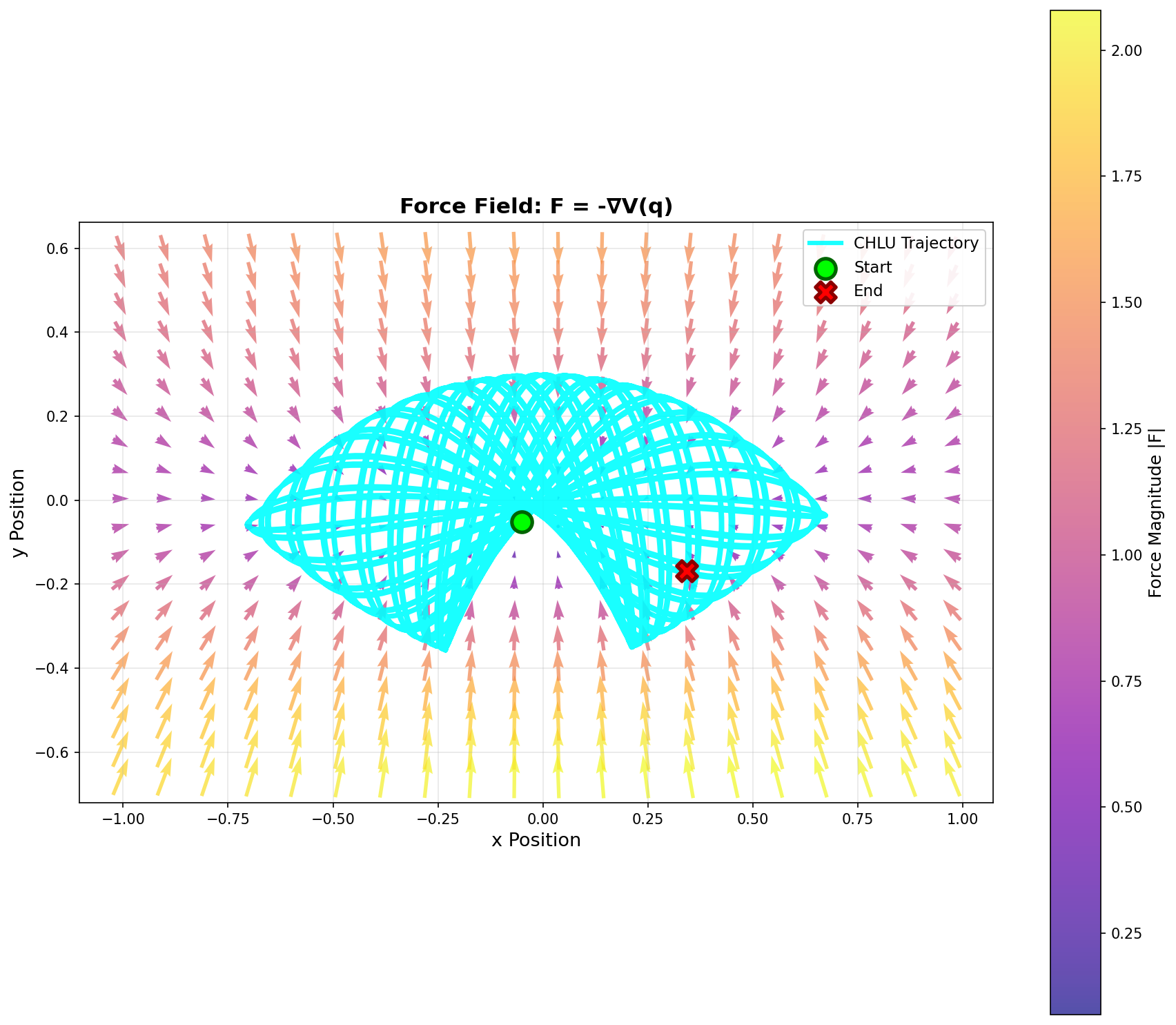}
\end{center}
\caption{The learned lemniscate potential energy plotted as $2D$, $3D$ heatmaps (top left, right) as well as a force field (bottom). The output trajectory is overlayed on all $3$ for visual comparison.}
\label{fig:exp1_potentialSurface}
\end{figure}

We probed the learned potential energy, $V_\theta(q)$ to find interpretable plots, Fig.\ref{fig:exp1_potentialSurface}, that show the stable attractor we intended to learn.

\subsection{Perturbed Sine Wave}
\begin{figure}[ht]
\begin{center}
\includegraphics[width=0.7\linewidth]{figs/exp2_kinetic_energy_newtonian.png} \includegraphics[width=0.7\linewidth]{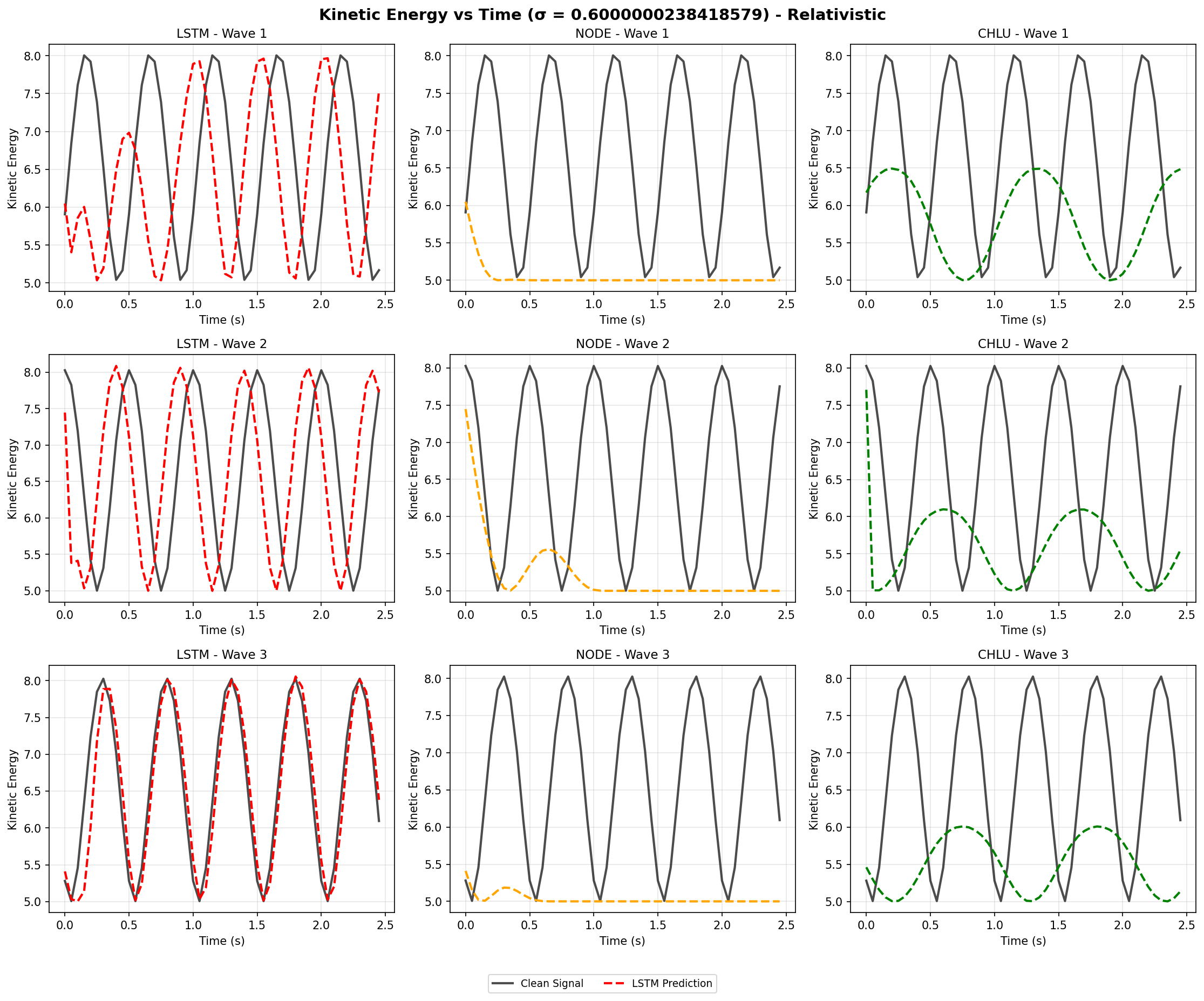}
\end{center}
\caption{Kinetic energies calculated from the output state $q$ (top 3 rows), compared to the kinetic energies calculated from output state $p$ (bottom 3 rows) for the same $3$ sine waves with the same inital perturbed conditions for all models (left to right: LSTM, NODE, CHLU).}
\label{fig:exp2_model}
\end{figure}

We showed the kinetic energy calculated from the output state $q$ for all 3 models in Fig.\ref{fig:exp2_noise}. Here we show the same plot on the top 3 rows of Fig.\ref{fig:exp2_model}. The bottom 3 rows, however, are the kinetic energies for the same inference steps, but this time calculated using the output state $p$. Although the LSTM produces near perfect kinetic energy based on $p$, the LSTM fails to learn the fundamental correlation between $\dot{q}$ and $p$, and as a result produces 2 different initial kinetic energies for a single system. This further shows that LSTMs are handicapped by design when the goal is to learn physically consistent internal models since its only objective is to minimize next-step prediction error. The CHLU on the other hand collapses to a phase-shifted, magnitude conserved sine wave, since it is inherently biased to conserve learned energy states in the form of stable attractors. We discuss this as a failure mode in App.\ref{sec:failuremodes}. It is important to note that we are not comparing direct performance here, but showcasing the inductive biases of our model in contrast to the baselines.

\begin{figure}[ht]
\begin{center}
\includegraphics[width=0.8\linewidth]{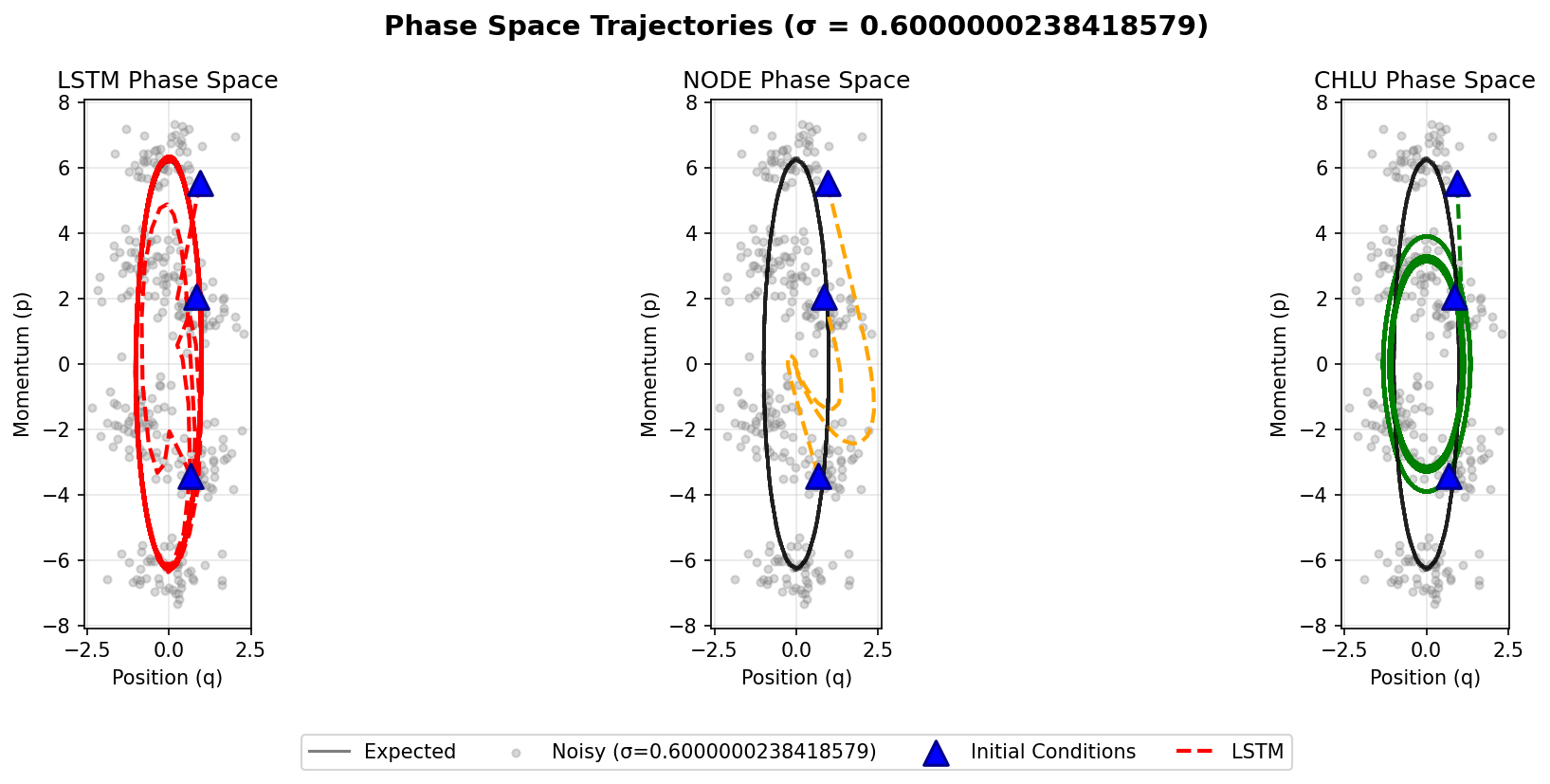} 
\end{center}
\caption{Phase space plots corresponding to the kinetic energy plots in Figs.\ref{fig:exp2_noise},\ref{fig:exp2_model}, with the blue triangles representing the perturbed initial states and the grey ellipse showing the expected trajectory (left to right: LSTM, NODE, CHLU).}
\label{fig:exp2_phasespace}
\end{figure}

The phase space plots for the produced sine waves themselves are shown in Fig.\ref{fig:exp2_phasespace}. The LSTM locks onto the perfect loop over time, but the initial few steps from the perturbed conditions show that the model has discontinuous phase space trajectories that are non-physical. The CHLU locks onto the nearest sine wave trajectory based on the initial noise dissipation that includes partial information dissipation as well (discussed in App.\ref{sec:failuremodes}).

\subsection{MNIST Generation}
\begin{figure}[ht]
\begin{center}
\includegraphics[width=0.7\linewidth]{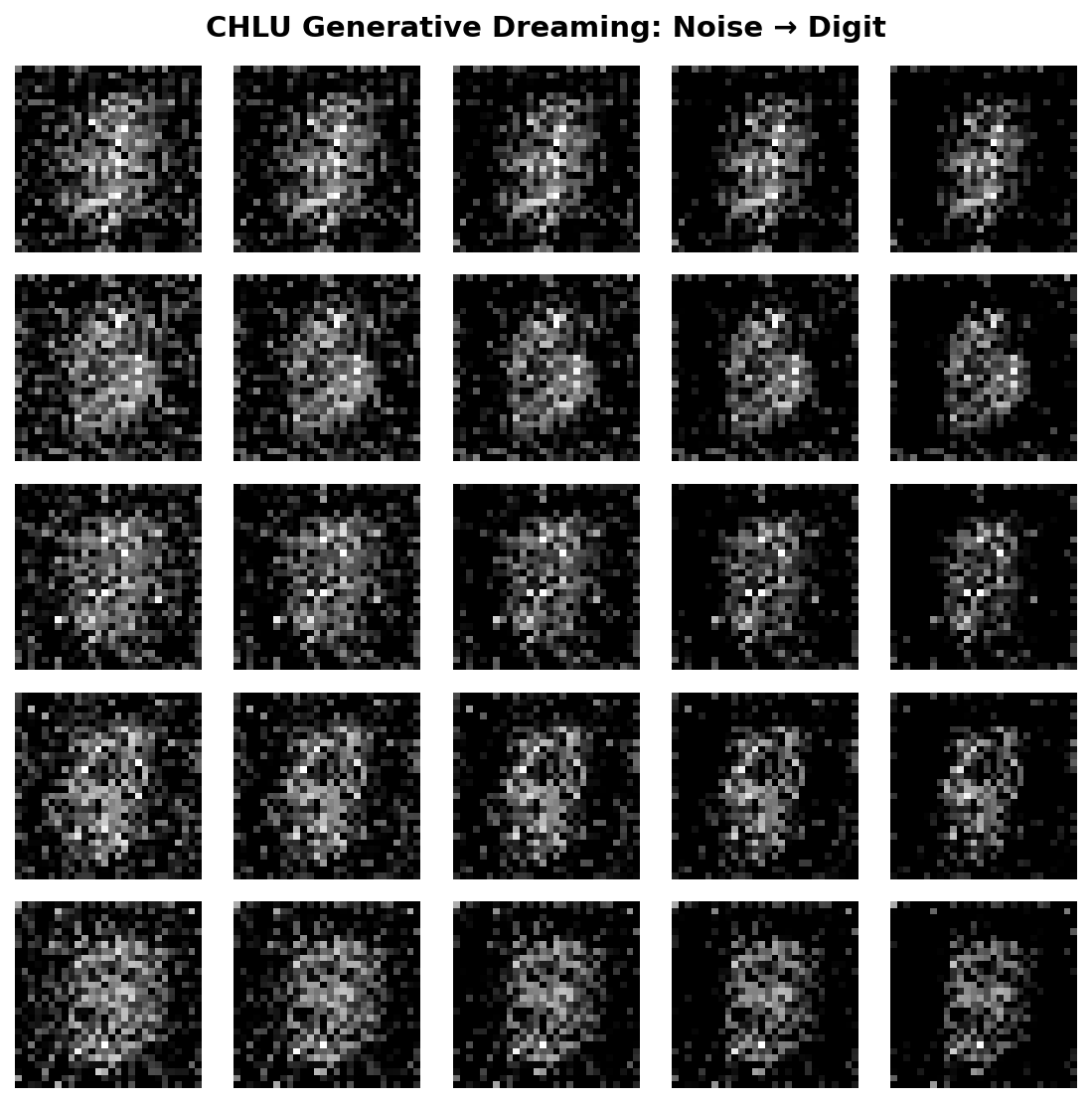} 
\end{center}
\caption{MNIST evolution from step $200-1000$ from left to right, at intervals of $200$ steps, for $5$ digits (rows).}
\label{fig:exp3_mnist_evol}
\end{figure}

Fig.\ref{fig:exp3_mnist} showed the generation results from the thermal-annealed mode (using and annealing both langevin temperature $\mathcal{T}$, and friction $\gamma$). The plot shown included a $tanh()$ function applied to the $1000$th generation step for better visualization. In Fig.\ref{fig:exp3_mnist_evol} we show the evolution of the first $5$ digits through steps $200-1000$, every $200$ steps.

\begin{figure}[h]
\begin{center}
\includegraphics[width=0.7\linewidth]{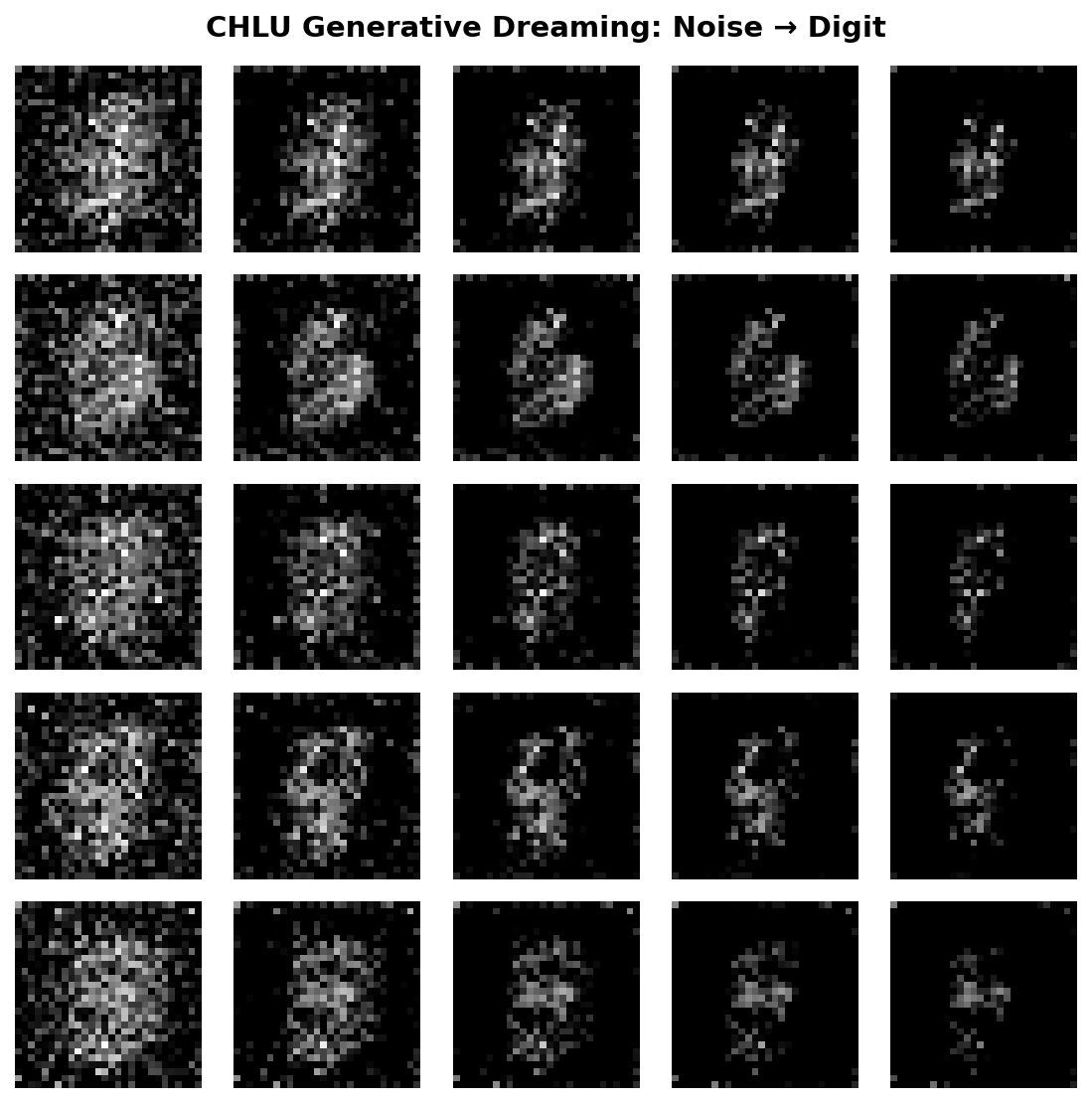} 
\end{center}
\caption{MNIST evolution from step $200-1000$ from left to right, at intervals of $200$ steps, for $5$ digits (rows), in the non-thermal conservative mode.}
\label{fig:exp3_mnist_ghost}
\end{figure}

Fig.\ref{fig:exp3_mnist_ghost} shows a generation following non-thermal, deterministic generation without friction ($\mathcal{T}$ and $\gamma$ set to $0$). Here the system settles into denoised digits faster, but the generation seems to drive all output pixels to the lowest values showing that the most stable state learned by the non-thermal deterministic values is a collapsed low energy representation where all pixels are set to their lowest values. In comparison the stochastic langevin mode causes the system to settle in non-collapsed states showing distinct digits.

\section{Limitations and Failure Modes:}
\label{sec:failuremodes}
It is crucial to note that a CHLU's strength, conservation, is also its primary limitation.
\begin{enumerate}
    \item \textbf{Hyper-Stability:} Because a CHLU is strictly conservative, it lacks a natural mechanism to "forget" noise. In purely dissipative systems (e.g., a damped pendulum), a conservative CHLU will oscillate forever unless the friction parameter is used, thereby giving up long-horizon stability. However, we argue that a solution to the memory-stability tradeoff is a combination of a Dissipative CHLU and a Symplectic RNN, giving design flexibility from both ends. 
    \item \textbf{Stiffness:} The Velocity Verlet integrator, while symplectic, can become unstable if the learned potential $V_\theta$ becomes extremely stiff (very high curvature), requiring adaptive time-stepping which breaks symplecticity.
\end{enumerate}

\section{Future Directions}
Future work will explore "Deep Symplectic Networks," stacking/connecting multiple CHLUs to model complex systems, including a comprehensive study of performance, biases and limitations.

We identify four key areas for future expansion:
\begin{itemize}
    \item \textbf{Lorentz Boosting (Lateral Energy):} A proposed mechanism to resolve ambiguity by applying a "Lorentz Boost" to the latent reference frame as a form of global attention. This goal is to mechanically cluster semantically related features without altering weights during inference.
    \item \textbf{Wormholes:} Implementing sparse non-local connections penalized by energy cost (rather than physical changes to space-time topology) to allow ``tunneling" across the causal diamond, facilitating long-range information transfer without $O(N^2)$ complexity. This allows for ad-hoc breaking of CHLU's internal conservation laws to provide sparse-non local attention, potentially a key requirement to reach comparable performance to modern deep learning algorithms on fundamnetally non-causal tasks.
    \item \textbf{Connected CHLU Netowrks:} Connecting multiple CHLUs to allow each unit to propagate learned information at scale.
    \item \textbf{Hierarchical Causality:} Stacking CHLUs where the ``Output Shell" of one unit becomes the ``Input Shell" (Boundary Condition) of the next, allowing for multi-scale physical modeling.
    \item \textbf{Shell Jumping:} An Adaptive Computation Time mechanism where the network projects the state to higher-dimensional manifolds ("Shells") when uncertainty is high.
    
\end{itemize}

\subsubsection*{Acknowledgments}
PJ's and MP's work was carried out as part of the \texttt{SMARTHEP} MSCA-ITN. \texttt{SMARTHEP} is funded by the European Union’s Horizon 2020 research and innovation programme, call H2020-MSCA-ITN-2020, under Grant Agreement n. 956086.

\end{document}